\documentclass{article}
\usepackage{spconf,amsmath,epsfig}

\usepackage[colorlinks,linkcolor=black]{hyperref}

\usepackage[normalem]{ulem}
\useunder{\uline}{\ul}{}
\usepackage{amsmath}
\usepackage{subfig}
\newcommand{\tabincell}[2]{\begin{tabular}{@{}#1@{}}#2\end{tabular}}
\begin{document}\sloppy

\def\x{{\mathbf x}}
\def\L{{\cal L}}

\title{ Locality-constrained Spatial Transformer Network\\ for Video Crowd Counting}
%

\renewcommand{\thefootnote}{\fnsymbol{footnote}}
\name{
Yanyan Fang \textsuperscript{a}  ,Biyun Zhan \textsuperscript{a}, Wandi Cai \textsuperscript{a}, Shenghua Gao \textsuperscript{b}, Bo Hu \textsuperscript{a,$*$}}
\address{ \textsuperscript{a}School of information and technology, Fudan University, \textsuperscript{b}Shanghaitech University\\
$\{$yyfang, byzhan15, wdcai15, bohu$\}$@fudan.edu.cn, gaoshh@shanghaitech.edu.cn}

\maketitle
\renewcommand{\thefootnote}{\fnsymbol{footnote}} 
\footnotetext[1]{Corresponding author.} 

\begin{abstract}
Compared with single image based crowd counting, video provides the spatial-temporal information of the crowd that would help improve the robustness of crowd counting. But translation, rotation and scaling of people lead to the change of density map of heads between neighbouring frames. Meanwhile, people walking in/out or being occluded in dynamic scenes leads to the change of head counts. To alleviate these issues in video crowd counting, a Locality-constrained Spatial Transformer Network (LSTN) is proposed. Specifically, we first leverage a Convolutional Neural Networks to estimate the density map for each frame. Then to relate the density maps between neighbouring frames, a Locality-constrained Spatial Transformer (LST) module is introduced to estimate the density map of next frame with that of current frame.
To facilitate the performance evaluation, a large-scale video crowd counting dataset is collected, which contains 15K frames with about 394K annotated heads captured from 13 different scenes. As far as we know, it is the largest video crowd counting dataset. Extensive experiments on our dataset and other crowd counting datasets validate the effectiveness of our LSTN for crowd counting. All our dataset are released in \url{ https://github.com/sweetyy83/Lstn_fdst_dataset.}

\end{abstract}
\begin{keywords}
Convolutional Neural Network; Locality-constrained Spatial Transformer Network; Video Crowd Counting
\end{keywords}
\vspace{-10pt}
\section{Introduction}

Crowd counting has been widely used in computer vision because of its potential applications in video surveillance, traffic control, and emergency management. However, most previous works \cite{FU201581}\cite{Zhang_2015_CVPR}\cite{7780439} focus on single image based crowd counting. In real applications, we have videos at hand, and usually the movement of crowd is predictable and consistent \cite{federico2017context-aware}. In this paper, we target at exploiting the spatial-temporal consistency among neighbouring frames for more robust video crowd counting.

Previous crowd counting methods can be roughly categorized into detection-based approaches and regression-based approaches. Detection based approaches count crowd by detecting heads or pedestrians, but these approaches usually fail to detect tiny \cite{dalal2005histograms} or occluded \cite{tuzel2008pedestrian} heads/bodies which are very common in real scenarios. Thus regression-based approaches are more commonly used. Recently, in light of the success of Convolutional Neural Networks (CNN) for image classification, it also has been introduced to crowd counting, where CNN is used to learn a mapping from an input image to its corresponding density map. To leverage the spatial-temporal consistency among neighbouring frames for more accurate density maps in videos, LSTM \cite{8237658} or ConvLSTM \cite{xiong2017spatiotemporal} based approaches have been proposed which accumulate features of all history frames with LSTM or ConvLSTM for density map estimation. These approaches have demonstrated their effectiveness for video crowd counting, but they leverage history information in an implicit way, and as people walk in/out or are occluded, the identities of the crowd in the history frame may be totally different from the ones in current frame. Consequently, the features from these history may even hurt the density map estimation of current frame.

Rather than using LSTM or ConvLSTM to implicitly model the spatial-temporal dependencies in videos, in this paper, we propose to leverage a Locality-constrained Spatial Transformer (LST) module to explicitly model the spatial-temporal correlation between neighbouring frames. Specifically, on one hand, given the same population of the crowd, previous work \cite{federico2017context-aware} has shown that the trajectories of crowd can be well predicted. But because of the change of perspective, distance, rotation, and lighting, the appearance of the same person may visually change a lot, and thus it sometimes may be not easy to directly re-identify the people in two adjacent frames. But density map ignores the appearances of the people and is only related to the location of heads. Since people's trajectories are predictable, the density map of one frame probably can be warped from that of its previous frame with some transformations, including scaling and translation caused by people walking away from or towards camera, rotation caused by the motion of camera, \emph{etc.}. On the other hand, for videos, some people walk in/out of the imaging range of camera or are occluded. In these cases, it is infeasible to estimate the density maps for those people from previous frames. By taking all these factors together, in our LST, rather than warping the density map for the whole frame, we propose to divide each frame into blocks. Given two blocks with the same location but from two neighbouring frames, we use their similarity to weight the difference between the ground-truth density map of the block and the one warped from the density map of the other block. If these two blocks are similar, they probably correspond to the same population, then the difference between ground-truth density map and warped density map should be smaller. If someone walks in/out or is occluded, then we allow the warped density map from previous frame to be slightly different from the ground-truth. Further, since only the spatial-temporal dependencies between neighbouring frames are used, our model can get rid of the effect of irrelevant history frames in density map estimation. Experiments validate the effectiveness of our model for video crowd counting.

A large-scale dataset with multiple scenes is desirable for video crowd counting. But most existing datasets are too small and with only a few scenes. For example, the WorldExpo'10 dataset, which is the largest one in previous works, only contains 5 scenes. Thus we propose to build a new large-scale video crowd counting dataset named Fudan-ShanghaiTech (FDST) with more scenes. Specifically, FDST dataset contains 15,000 frames with 394,081 annotated heads captured from 13 difference scenes, including shopping malls, squares, hospitals, \emph{etc.}. The dataset is much larger than the WorldExpo'10 dataset, which only contains 3980 frames with 199,923 annotated heads. Further, we provide the frame-wise annotation while WordExPo'10 only provides the annotation for every 30 seconds. Therefore FDST dataset is more suitable for video crowd counting evaluation. 

The main contributions of our work can be summarized as follows: i) we propose a Locality-constrained Spatial Transformer Network (LSTN), which explicitly models the spatial-temporal dependencies between neighbouring frames to facilitate the video crowd counting; ii) we collect a large-scale video crowd counting dataset with frame-wise ground-truth annotation, which would facilitate the performance evaluation in video crowd counting; iii) extensive experiments validate the effectiveness of our model for video crowd counting.


\vspace{-10pt}
 \section{Related work}
Since our work is related to deep learning based crowd counting, here we only briefly discuss recent works on deep learning based crowd counting.

\textbf{Crowd counting for single image.} Recent works \cite{7780439}\cite{Sam_2017_CVPR}\cite{li2018csrnet} have shown the effectiveness of CNN for density map estimation in single image crowd counting. To improve the robustness of crowd counting for areas with different head sizes and densities, different network architectures have been proposed, including MCNN \cite{7780439}, Hydra CNN \cite{onoro2016towards}, Switch-CNN \cite{Sam_2017_CVPR}, CSRNet \cite{li2018csrnet}, which basically leverages networks with different local receptive fields for density maps estimation. Further, recently, people also propose to leverage detection \cite{liu2018decidenet} or localization \cite{idrees2018composition} tasks to assist the crowd counting task. But these single image crowd counting methods may lead to inconsistent head counts for neighbouring frames in video crowd counting.

\textbf{Video crowd counting.} Most previous works focus on single image crowd counting and there are only a few works on video crowd counting. Recently, Xiong \emph{et al.} \cite{xiong2017spatiotemporal} propose to leverage ConvLSTM to integrate history features and features of current frame for video crowd counting, which has shown its effectiveness for video crowd counting. Further, Zhang \emph{et al.} \cite{8237658} also propose to use LSTM for vehicle counting in videos. However, all these LSTM based methods may be affected by those irrelevant history, and do not explicitly consider the spatial-temporal dependencies in videos, whereas our solution models such dependencies in neighbouring frames with LST explicitly. Thus our solution is more straightforward.

\textbf{Spatial transformer network (STN).} Recently, Jaderberg \emph{et al.} \cite{jaderberg2015spatial}introduce a differentiable Spatial Transformer (ST) module which is capable to model the spatial transformation between input and output. Such ST module can be easily plugged into many existing networks and trained in an end-to-end manner, and has shown its effectiveness for face alignment \cite{chen2016supervised}\cite{zhong2017toward} and face recognition \cite{wu2017recursive}. Further, it also has been applied for density map estimation in a coarse-to-fine based single image crowd counting framework \cite{liu2018crowd}. But different from \cite{liu2018crowd}, we propose to leverage ST to relate density maps between neighbouring frames for video crowd counting.

\begin{figure}[!h]
    \centering
	\includegraphics[scale=0.52]{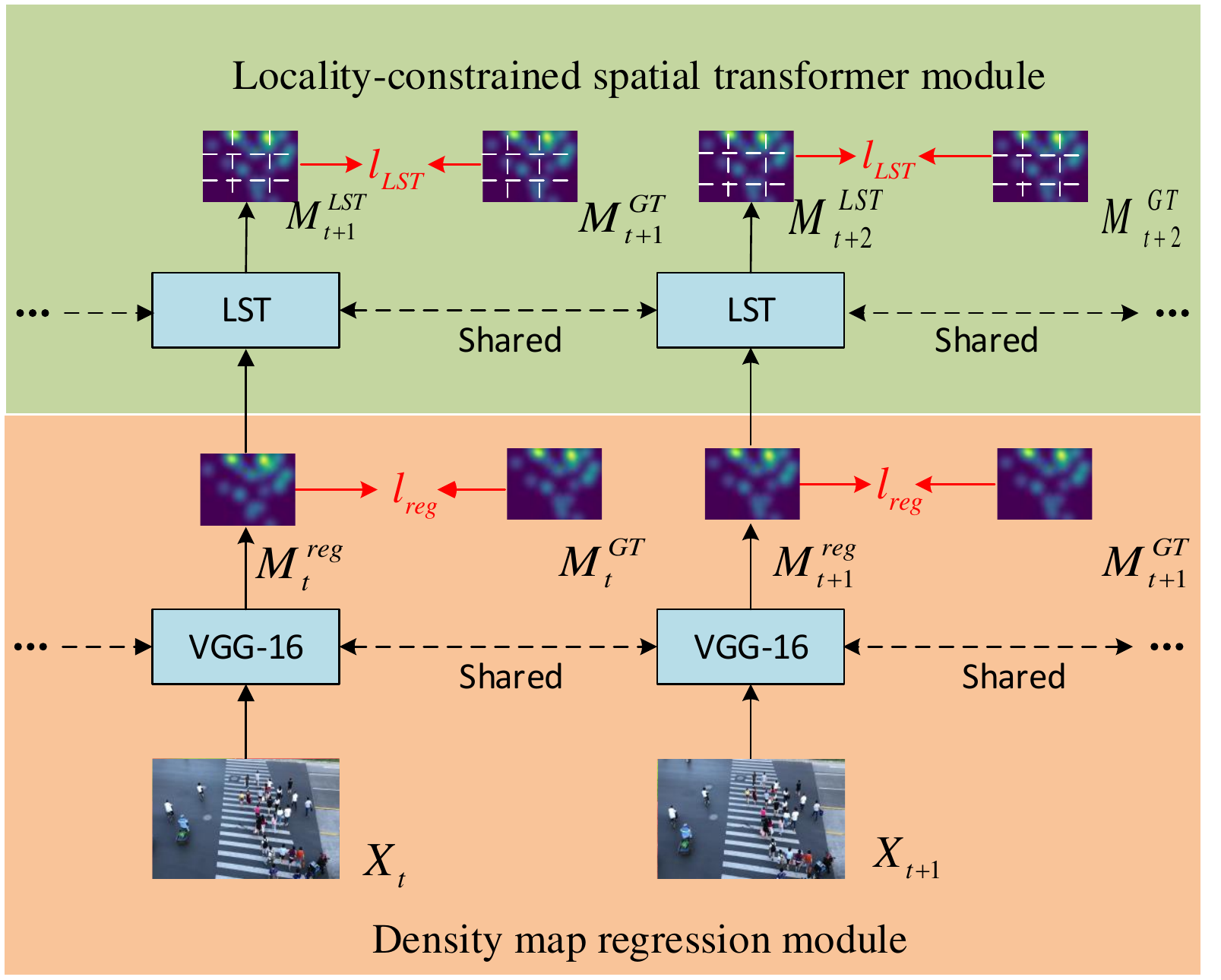}
	\caption{The structure of the LSTN module for video crowd counting.}
	\label{module}
\end{figure}


\vspace{-10pt}
\section{Our Approach}

Our network architecture is shown in Fig. \ref{module}. It consists of two modules: density map regression module and Locality-constrained Spatial Transformer (LST) module. The density map regression module takes each frame as input and estimates its corresponding density map, and then the LST module takes the estimated density map as input to predict the density map of next frame.

\subsection{Density map regression module}
Density map generation is very important for the performance of density map based crowd counting. Given one frame with $N$ heads, if the $i^{th}$ head is centered at $p_i$, we represent it as a delta function $\delta(p-p_i)$. Hence the ground-truth density map of this frame can be calculated as follows:
\begin{equation}
\mathcal{M} = \sum\limits_{i=1}^N \delta(p-p_i) * G_\sigma(p).
\label{densityMap}
\end{equation}
Here $G_\sigma(p)$ is a 2D Gaussian kernel with variance $\sigma$:
\begin{equation}
G_\sigma(p) = \frac{1}{2 \pi \sigma^2}e^{-\frac{(x^2+y^2)}{2\sigma ^2}}
\label{Gaussian}
\end{equation}
In other words, if a pixel is near the annotated point, it has higher probability belonging to a head. Once the density map is defined, the density map regression module maps each frame to its corresponding density map. We denote the ground-truth density map of $t^{th}$ ($t=1, \ldots,T$) frame as $M_t^{GT}$, and denote the density map estimated by density map regression module as $M_t^{reg}$. Then the objective of density map regression module can be written as follows:
\begin{equation}
\ell_{\text{reg}}=\frac{1}{2T}\sum_{t=1}^T\quad\|M_t^{reg}-M_t^{GT}\|^2
\end{equation}

In our implementation, we use VGG-16 network in our density map regression module.

\begin{table*}[!t]
     \caption{Details of some datasets: Num is the total number of frames; FPS is the number of frames per second; Max is the maximal crowd count in one frame; Min is the minimal crowd count in one frame; Ave is the average crowd count in one frame; Total is total number of labeled people.}
	\label{table:label}
	\centering
	\begin{tabular}{|c|c|c|c|c|c|c|c|c|}
		\hline
           \textbf{Dataset}	& \textbf{Resolution}&  \textbf{Num} & \textbf{FPS}	&  \textbf{Max}	&  \textbf{Min}&  \textbf{Ave}&  \textbf{Total}	\\
		\hline
           \hline
		\textbf{UCSD} & $238\times158$ &2000  	&10	&46&11  & 24.9 & 49,885\\
		\hline

		\textbf{Mall}   & $640\times 480$ &2000 &$\textless 2$&53&13 & 31.2  &62,316		\\
           \hline
		\textbf{WorldExpo} & $576\times 720$ &3980&50 &253& 1      &50.2	   	 &199,923     \\
           \hline
		\textbf{Ours}   & \tabincell{c}{$1920\times1080$ \\ $1280\times720$} &15000&30&57&9 &26.7  &394,081	 \\
          \hline
      	\end{tabular}
      \label{tab}	
\end{table*}

\subsection{LST module}
For the same population of crowd in videos, many previous works have shown that the trajectories of these people can be well predicted. Thus the density map of previous frame would help the density map prediction of current frame. However, in all video crowd counting datasets, the correspondence of people in neighbouring frames are not provided, which prevents directly learning a mapping from head coordinates in previous frame to those in current frame. Further, because of the change in perspective, distance, rotation, and lighting condition in neighbouring frames as well as occlusion, the appearance of the same person may visually change a lot, which makes directly re-identifying the person in two frames difficult. But density map ignores the appearances of the people and is only related to the location of heads. Now that people's trajectories are predictable, we can leverage the density map of previous frame to estimate the density map of current frame for the same group of people. Specifically, the deformation of the density map for the same group people in neighbouring frames includes scaling and translation if people walk away from or towards camera, or rotation if there exists some motion for camera, for example, caused by wind or vibration of ground.


Recent work \cite{liu2018crowd} has shown the effectiveness of spatial transformer (ST) module for learning the transform between input and output. Thus ST can be used to learn the mapping for the same group of people between the two neighbouring frames. However, in practice, people walk in/out the range of camera, and some people may be occluded, which restricts the application of ST. Thus, in this paper, we propose an LST, which is essentially a weighted ST for each image block. Specifically, we divide each frame into many blocks. Given two blocks with the same spatial coordinates but from two neighbouring frames, we use their similarity to weight the difference between the ground-truth density map of one block and the density map transformed from the other block. If these two blocks are similar, they probably correspond to the same population, then the difference between ground-truth density map and transformed density map should be smaller. If someone walks in/out or is occluded, then we allow the estimated density map to be slightly different from the ground-truth. By minimizing such difference over all blocks and all frames, the dependencies between neighbouring frames can be exploited for video crowd counting.

We denote the mapping function of LST module as $f_{LST}$ which takes the estimated density map of the $t^{th}$ frame as input to estimate the density map of the $(t+1)^{th}$ frame. We use $M_{t+1}^{LST}$ to denote the density map of the $(t+1)^{th}$ frame estimated by LST. Then
\begin{equation}
M_{t+1}^{LST}= f_{LST}(M_t^{reg};A_\theta)
\end{equation}
\begin{equation}
\begin{split}
{\left[ \begin{array}{c}
x_i^s \\
y_i^s
\end{array}
\right ]} &=\Gamma_\theta(G_i)=A_\theta {\left[ \begin{array}{c}
x_i^t \\
y_i^t \\
1
\end{array}
\right ]}\\
 &={\left[ \begin{array}{ccc}
\theta_{11} & \theta_{12} & \theta_{13}\\
\theta_{21}& \theta_{22} & \theta_{23} \\
\end{array}
\right ]} {\left[ \begin{array}{c}
x_i^t \\
y_i^t \\
1
\end{array}
\right ]}
\end{split}
\end{equation}
where ($x_i^t$ ,$y_i^t$) are the target coordinates of the sampling grid $\Gamma_\theta $ in the output density maps, ($x_i^s$ ,$y_i^s$) are the source coordinates in the input density maps that define the sample points, and $A_\theta$ denotes the transformation matrix \cite{jaderberg2015spatial}.


We evenly divide each frame $I_{t}$, $M_{t+1}^{GT}$ and $M_{t+1}^{LST}$ into $H\times W$ blocks, and use $I_{t}(i,j)$, $M_{t+1}^{GT}(i,j)$ and $M_{t+1}^{LST}(i,j)$ to denote the block in the $j^{th}$ column and the $i^{th}$ row for the $t^{th}$ frame, its ground-truth density map and density map estimated by LST. Then the objective of LST can be written as follows.
\begin{equation}
\begin{split}
\ell_{\text{LST}} = \frac{1}{2T} \sum^{T-1}_{t=1}\sum_{\substack{1\le i \le H \\1\le j \le W}} &S(I_{t}(i,j),I_{t+1}(i,j))\\
                        &\times\|M_{t+1}^{LST}(i,j)- M_{t+1}^{GT}(i,j)\|^2_2
\end{split}
\end{equation}
where $S(I_{t}(i,j),I_{t+1}(i,j))$ denotes the similarity between the corresponding temporal neighbouring blocks, which can be measured as follows
\begin{equation}
\centering
S(I_{t}(i,j),I_{t+1}(i,j))=\exp(-\frac{\|I_{t}(i,j)-I_{t+1}(i,j)\|^2_2}{2\beta^2}) .
\end{equation}

\vspace{-3pt}
\subsection{Loss function}
We combine the losses of the density map regression module and that of the LST module, and arrive at the following objective function
\begin{equation}
\ell = \ell_{\text{reg}}+\lambda\ell_{\text{LST}},
\end{equation}
where $\lambda$ is a weight used to balance $\ell_{\text{reg}}$ and $\ell_{\text{LST}}$.

\par In the training process, an Adam optimizer is used with a learning rate at 1e-8 on our dataset. To reduce over-fitting, we adopt the batch-normalization, and the batch-size is 5.

Once our network is trained, in the testing phase, we can directly estimate the density map of each frame and integrate the density map to get the estimated head counts.

\subsection{Implementation details}
The variance in gaussian based density map generation $\gamma=3$, and the $\beta$ used in similarity measurement is 30 on FDST dataset. We resize all frames to $640\times360$ pixels. We first pretrain density map regression module, then we fine-tune the whole network by fix the first 10 layers in VGG-16. For the number of blocks, we fix $W=2$ on all datasets. On the Mall dataset and our dataset, we fix $H=1$, and $H=2$ on the UCSD dataset. We set $\lambda=0.001$ on FDST dataset\footnote[2]{Because the ground-truth are annotated 2fps on Expo'10 and ROI's are also marked, therefore the population of two neighbouring frames change a lot. Thus this dataset is not suitable for performance evaluation of our method.}.

\section{Experiments }

\subsection{Evaluation metric}
Following work \cite{tota2015counting}, we adopt both the mean absolute error (MAE) and the mean squared error (MSE) as metrics to evaluate the performance of different methods, which are defined as follows:
\begin{equation}
\centering
MAE = \frac{1}{T}\sum_{i=1}^T |z_i - \hat{z_i}|,  MSE = \sqrt{\frac{1}{T}\sum_{i=1}^T (z_i-\hat{z_i})^2}
\end{equation}
where $T$ is the total number of frames of all testing video sequences, $z_i$ and $ \hat{z_i}$ are the actual number of people and estimated number of people in the $i^{th}$ frame respectively.

\subsection{Fudan-ShanghaiTech Video Crowd counting dataset}
Existing video crowd counting datasets are too small in terms of number of both frames as well as scenes. Hence, we introduce a new large-scale video crowd counting dataset.
Specifically, we collected 100 videos captured from 13 different scenes, and FDST dataset contains 150,000 frames, with a total of 394,081 annotated heads.  It takes more than 400 hours to annotate FDST dataset. As far as we know, this dataset is the largest video crowd counting dataset. Table.\ref{tab} shows the statistics of our dataset and other relevant datasets.

\begin{figure}[!h]
    \centering
	\includegraphics[scale=0.6]{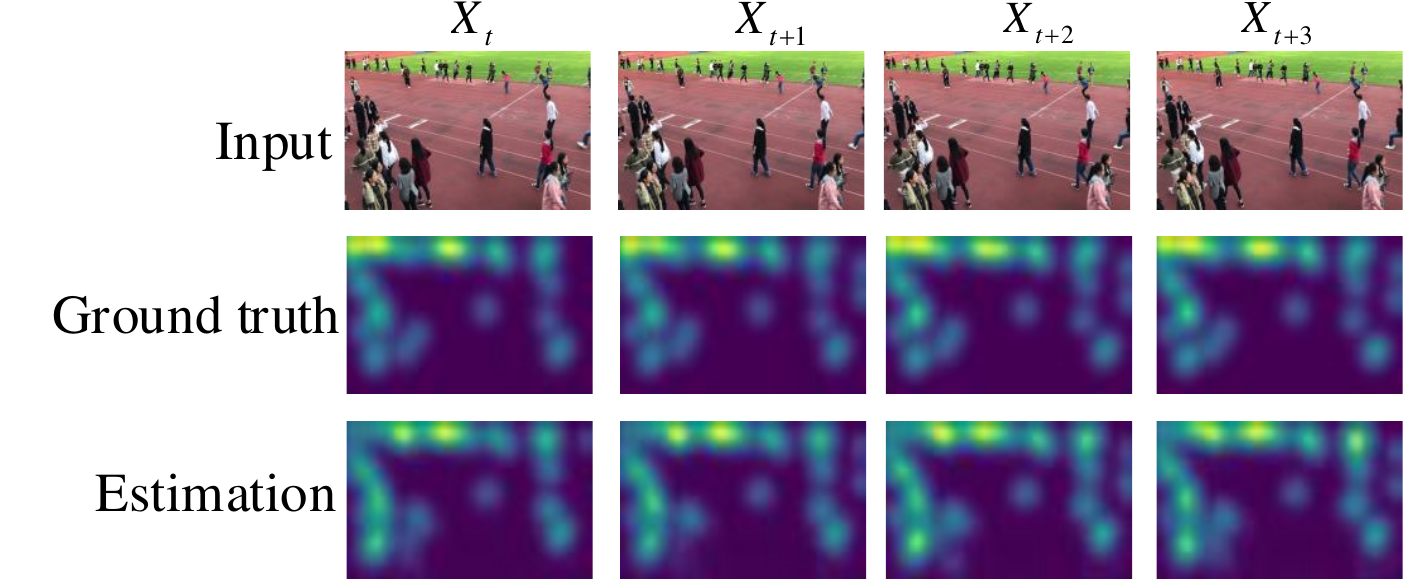}
	\caption{The density maps estimated by our method on our dataset. }
	\label{visual}	
\end{figure}

The training set of FDST dataset consists of 60 videos, 9000 frames and the testing set contains the remaining 40 videos, 6000 frames. We compare our method with MCNN \cite{7780439} which achieves state-of-the-art performance for single image crowd counting, ConvLSTM \cite{xiong2017spatiotemporal} which is state-of-the-art video crowd counting method. We also report the performance of our method without LST. All results are shown in Table. \ref{our}. We can see that our method achieves the best performance. Further the improvement of our method compared with the one without LST shows the effectiveness of LST. It is worth noting that because there are many scenes in our dataset, and it is not easy to train the ConvLSTM, therefore the performance of ConvLSTM is even worse than single image based method. We also show the density map estimated by our LSTN in Fig. \ref{visual}.

\begin{table}[h]
     \caption{Results of different methods on our dataset.}
	\label{table:label}
	\centering
	\begin{tabular}{|c|c|c|}
	\hline
			\textbf{Method}	& \textbf{MAE}	&  \textbf{MSE}	\\
	
	\hline
			{MCNN \cite{7780439}} &3.77	&	4.88\\
	\hline
			{ConvLSTM \cite{xiong2017spatiotemporal}}	&4.48	&5.82	\\
    \hline {Ours without LST} &3.87 &5.16\\
	\hline
			{Our Method}		&\textbf{3.35}	&\textbf{4.45}	\\
	\hline
	\end{tabular}
	\label{our}	
\end{table}

\vspace{-15pt}
\subsection{The UCSD dataset}
We also evaluate our method with the UCSD dataset \cite{4587569}, which contains 2000 frames captured by surveillance cameras in the UCSD campus. The resolution of frames is $238\times158$ pixels and the rate of frame is 10 fps. The number of person in each frame varies from 11 to 46. By following the same setting with \cite{4587569}, we use frames from 601 to 1400 as training data, and the remaining 1200 frames as testing data.

Following \cite{li2018csrnet}, we use bilinear interpolation to resize each frame into $952\times 632$. Table. \ref{ucsd} shows the accuracy of different methods on this dataset. We can see that our method also outperforms ConvLSTM based method on this dataset.
\begin{table}[h]
     \caption{Results of different methods on the UCSD dataset.}
	\label{table:label}
	\centering
	\begin{tabular}{|c|c|c|}
	\hline
			\textbf{Method}	& \textbf{MAE}	&  \textbf{MSE}	\\
	\hline
	\hline
			{Kernel Ridge Regression \cite{4270130}}		&2.16	&7.45 \\
	\hline
			{Ridge Regression \cite{Chen_featuremining}}				&2.25	&7.82 \\
	\hline
			{Gaussian Process Regression \cite{4587569}}	&2.24	&7.97 \\
	\hline
			{Cumulative Attribute Regression \cite{6619163}}	&2.07 	&6.86 \\
	\hline
			{Zhang \emph{et al} \cite{Zhang_2015_CVPR}}				&1.60 	&3.31 \\
	\hline
			{MCNN \cite{7780439}}						&\textbf{1.07}	&\textbf{1.35} \\
	\hline
			{Switch-CNN \cite{Sam_2017_CVPR}}					&1.62	&2.10 \\
	\hline
			{CSRNet \cite{li2018csrnet}}						&1.16	&1.47 \\
	\hline
			{FCN-rLSTM \cite{8237658}}					&1.54	&3.02 \\
	\hline
			{ConvLSTM \cite{xiong2017spatiotemporal}}					&1.30	&1.79 \\
	\hline
			{Bidirectional ConvLSTM \cite{xiong2017spatiotemporal}}			&1.13 	&1.43 \\
	\hline
			{Our Method}					&\textbf{1.07}	&1.39 \\
	\hline
	\end{tabular}
	\label{ucsd}		
\end{table}

\subsection{The Mall dataset}
\par The Mall dataset is captured in a shopping mall with a surveillance camera \cite{Chen_featuremining}. This video-based dataset consists of 2000 frames in the dimension of $640\times480$ pixels, with over 60,000 labeled pedestrians. Region of Interest (ROI) and perspective map are also provided. According to the train-test setting in \cite{Chen_featuremining}, we use the first 800 frames for training and the remaining 1200 frames for testing.
The performance of different methods are shown in Table. \ref{mall}, our model also achieves state-of-the-art performance in terms of both MAE and MSE.
\begin{table}[h]
     \caption{Results of different methods on the Mall dataset.}
	\label{table:label}
	\centering
	\begin{tabular}{|c|c|c|}
	\hline
			\textbf{Method}	& \textbf{MAE}	&  \textbf{MSE}	\\
	\hline
	\hline
			{Kernel Ridge Regression \cite{4270130}}		&3.51	&18.10 \\
	\hline
			{Ridge Regression \cite{Chen_featuremining}}				&3.59 	&19.00 \\
	\hline
			{Gaussian Process Regression \cite{4587569}}	&3.72	&20.10 \\
	\hline
			{Cumulative Attribute Regression \cite{6619163}}	&3.43 	&17.70 \\
	\hline
			{COUNT Forest \cite{7410729}}				&2.50 	&10.00 \\
	\hline
			{ConvLSTM \cite{xiong2017spatiotemporal}}					&2.24	&8.50 \\
	\hline
			{Bidirectional ConvLSTM \cite{xiong2017spatiotemporal}}			&2.10 	&7.60	\\
	\hline
			{Our Method}					&\textbf{2.00}	&\textbf{2.50} \\
	\hline
	\end{tabular}
	\label{mall}		
\end{table}


\subsection{The importance of similarity term in LST}
In our LSTN, we use the similarity between temporal neighbouring blocks to weight the difference between the warped density map and its ground-truth. The underlying assumption is that if two blocks are similar, then the population within these two blocks probably correspond to the same group of people and then spatial transformer works well. But if the similarity is lower, which means people walk in/out or are occluded, then it is less possible to infer the density map of block in the temporal neighbouring frame.
We compare the results with/without similarity term on UCSD, Mall, FDST dataset, and the results are shown in Table.\ref{similarity}. We can see that similarity term always boots the performance of video crowd counting, which validates our assumption.

\begin{table}[h]
     \caption{Comparing the performance of  LSTN with/without similarity term on UCSD, Mall and our dataset.}
	\label{table:label}
	\centering
	\begin{tabular}{|c|c|c|c|c|}
	\hline
           \                            	&\multicolumn{2}{c|}{ \textbf{With}} &\multicolumn{2}{c|}{\textbf{Without}}\\
		\hline
           \textbf{Method}	& \textbf{MAE}&  \textbf{MSE} & \textbf{MAE}&  \textbf{MSE}\\
		\hline
           \hline
		\textbf{UCSD}   	&1.07&1.39  &1.11  & 1.41\\
		\hline

		\textbf{Mall}    &2.00&2.50 &2.18   &2.70	\\

	    \hline
		\textbf{Ours}  &3.35&4.45 &3.81  &5.10 \\
          \hline
      	\end{tabular}
      \label{similarity}	
\end{table}

\vspace{-15pt}
\section{Conclusion}
In this paper, a Locality-constrained Spatial Transformer Network (LSTN) is proposed to explicitly relate the density maps of neighbouring frames for video crowd counting. Specifically, we first leverage a density map regression module to estimate the density map of each frame. Considering that people may walk in/out or are occluded, we divide each frame into blocks, and use the similarity between two temporal neighbouring blocks to weight the difference between the ground-truth density map and the estimated one from the other block. We further build a large-scale video crowd counting dataset for performance evaluation, and as far as we know, FDST dataset is the larget video crowd counting dataset in terms of the number of both scenes and frames. Extensive experiments validate the effectiveness of our LSTN for video crowd counting.


\vspace{-15pt}
{


\begin{thebibliography}{10}

\bibitem{FU201581}
M.~Fu, P.~Xu, X.~Li, Q.Liu, M.Ye, and C.Zhu,
\newblock ``Fast crowd density estimation with convolutional neural networks,''
\newblock {\em Engineering Applications of Artificial Intelligence}, pp. 81 --
  88, 2015.

\bibitem{Zhang_2015_CVPR}
Cong Zhang, Hongsheng Li, Xiaogang Wang, and Xiaokang Yang,
\newblock ``Cross-scene crowd counting via deep convolutional neural
  networks,''
\newblock in {\em CVPR}, June 2015.

\bibitem{7780439}
Y.~Zhang, D.~Zhou, S.~Chen, S.~Gao, and Y.~Ma,
\newblock ``Single-image crowd counting via multi-column convolutional neural
  network,''
\newblock in {\em CVPR}, June 2016, pp. 589--597.

\bibitem{federico2017context-aware}
B.~Federico, L.~Giuseppe, Ballan L, and A.~Bimbo,
\newblock ``Context-aware trajectory prediction,''
\newblock {\em international conference on pattern recognition}, 2017.

\bibitem{dalal2005histograms}
N.~Dalal and B.~Triggs,
\newblock ``Histograms of oriented gradients for human detection,''
\newblock pp. 886--893, 2005.

\bibitem{tuzel2008pedestrian}
Oncel Tuzel, Fatih Porikli, and Peter Meer,
\newblock ``Pedestrian detection via classification on riemannian manifolds,''
\newblock {\em TPAMI}, vol. 30, no. 10, pp. 1713--1727, 2008.

\bibitem{8237658}
S.~Zhang, G.~Wu, J.~P. Costeira, and J.~M.~F. Moura,
\newblock ``Fcn-rlstm: Deep spatio-temporal neural networks for vehicle
  counting in city cameras,''
\newblock in {\em ICCV}, Oct 2017, pp. 3687--3696.

\bibitem{xiong2017spatiotemporal}
X.~Feng, X.~Shi, and D.~Yeung,
\newblock ``Spatiotemporal modeling for crowd counting in videos,''
\newblock in {\em ICCV}. IEEE, 2017, pp. 5161--5169.

\bibitem{Sam_2017_CVPR}
Deepak Babu~Sam, Shiv Surya, and R.~Venkatesh~Babu,
\newblock ``Switching convolutional neural network for crowd counting,''
\newblock in {\em CVPR}, July 2017.

\bibitem{li2018csrnet}
Y.~Li, X.~Zhang, and D.~Chen,
\newblock ``Csrnet: Dilated convolutional neural networks for understanding the
  highly congested scenes,''
\newblock in {\em CVPR}, 2018, pp. 1091--1100.

\bibitem{onoro2016towards}
Daniel D.~Onoro-Rubio and R.~L{\'o}pez-Sastre,
\newblock ``Towards perspective-free object counting with deep learning,''
\newblock in {\em ECCV}. Springer, 2016, pp. 615--629.

\bibitem{liu2018decidenet}
J.~Liu, C.~Gao, D.~Meng, and A.~Hauptmann,
\newblock ``Decidenet: counting varying density crowds through attention guided
  detection and density estimation,''
\newblock in {\em CVPR}, 2018, pp. 5197--5206.

\bibitem{idrees2018composition}
M.~Tayyab H.~Idrees, K.~Athrey, D.~Zhang, S.~Almaadeed, N.~Rajpoot, and
  M.~Shah,
\newblock ``Composition loss for counting, density map estimation and
  localization in dense crowds.,''
\newblock {\em arXiv: Computer Vision and Pattern Recognition}, 2018.

\bibitem{jaderberg2015spatial}
Max Jaderberg, Karen Simonyan, Andrew Zisserman, et~al.,
\newblock ``Spatial transformer networks,''
\newblock in {\em Advances in neural information processing systems}, 2015, pp.
  2017--2025.

\bibitem{chen2016supervised}
Dong Chen, Gang Hua, Fang Wen, and Jian Sun,
\newblock ``Supervised transformer network for efficient face detection,''
\newblock in {\em ECCV}. Springer, 2016, pp. 122--138.

\bibitem{zhong2017toward}
Yuanyi Zhong, Jiansheng Chen, and Bo~Huang,
\newblock ``Toward end-to-end face recognition through alignment learning,''
\newblock {\em IEEE signal processing letters}, vol. 24, no. 8, pp. 1213--1217,
  2017.

\bibitem{wu2017recursive}
Wanglong Wu, Meina Kan, Xin Liu, Yi~Yang, Shiguang Shan, and Xilin Chen,
\newblock ``Recursive spatial transformer (rest) for alignment-free face
  recognition,''
\newblock in {\em CVPR}, 2017, pp. 3772--3780.

\bibitem{liu2018crowd}
Lingbo Liu, Hongjun Wang, Guanbin Li, Wanli Ouyang, and Liang Lin,
\newblock ``Crowd counting using deep recurrent spatial-aware network,''
\newblock {\em arXiv preprint arXiv:1807.00601}, 2018.

\bibitem{tota2015counting}
Karunya Tota and Haroon Idrees,
\newblock ``Counting in dense crowds using deep features,'' 2015.

\bibitem{4587569}
A.~B. Chan, Zhang-Sheng~John Liang, and N.~Vasconcelos,
\newblock ``Privacy preserving crowd monitoring: Counting people without people
  models or tracking,''
\newblock in {\em CVPR}, June 2008, pp. 1--7.

\bibitem{4270130}
S.~An, W.~Liu, and S.~Venkatesh,
\newblock ``Face recognition using kernel ridge regression,''
\newblock in {\em CVPR}, June 2007, pp. 1--7.

\bibitem{Chen_featuremining}
Ke~Chen, Chen~Change Loy, Shaogang Gong, and Tao Xiang,
\newblock ``Feature mining for localised crowd counting,''
\newblock in {\em In BMVC}.

\bibitem{6619163}
K.~Chen, S.~Gong, T.~Xiang, and C.~C. Loy,
\newblock ``Cumulative attribute space for age and crowd density estimation,''
\newblock in {\em CVPR}, June 2013, pp. 2467--2474.

\bibitem{7410729}
V.~Pham, T.~Kozakaya, O.~Yamaguchi, and R.~Okada,
\newblock ``Count forest: Co-voting uncertain number of targets using random
  forest for crowd density estimation,''
\newblock in {\em ICCV}, Dec 2015, pp. 3253--3261.

\end{thebibliography}

}

\end{document}